\documentclass[sigconf,screen]{acmart}

\usepackage{amsmath}

\usepackage{algorithmic}
\usepackage{array}
\usepackage[caption=false,font=normalsize,labelfont=sf,textfont=sf]{subfig}
\usepackage{textcomp}
\usepackage{stfloats}
\usepackage{url}
\usepackage{verbatim}
\usepackage{graphicx}
\usepackage{balance}
\usepackage{hyperref}
\usepackage{multirow}
\usepackage{multicol}
\usepackage{enumitem}
\usepackage{arydshln} 
\usepackage{colortbl} 
\usepackage{pifont}

\AtBeginDocument{%
  \providecommand\BibTeX{{%
    \normalfont B\kern-0.5em{\scshape i\kern-0.25em b}\kern-0.8em\TeX}}}

\copyrightyear{2022} 
\acmYear{2022} 
\setcopyright{acmcopyright}\acmConference[MM '22]{Proceedings of the 30th ACM
International Conference on Multimedia}{October 10--14, 2022}{Lisboa, Portugal}
\acmBooktitle{Proceedings of the 30th ACM International Conference on Multimedia
(MM '22), October 10--14, 2022, Lisboa, Portugal}
\acmPrice{15.00}
\acmDOI{10.1145/3503161.3548358}
\acmISBN{978-1-4503-9203-7/22/10}

\newcommand{\model}{TransDIC}
\newcommand{\flows}{Two-Flow Encoder}

\definecolor{mygreen}{rgb}{0.09, 0.45, 0.27}
\definecolor{myred}{rgb}{1.0, 0.0, 0.0}
\definecolor{myyellow}{rgb}{1.0, 0.75, 0.0}
\definecolor{myblue}{rgb}{0.3, 0.55, 0.9}
\definecolor{mygray-bg}{gray}{0.9}

\renewcommand\appendix{\par
    \setcounter{section}{0}
    \setcounter{subsection}{0}
    \gdef\thesection{Appendix. \Alph{section}}}

\newcommand{\etal}{\textit{et al}. }
\newcommand{\ie}{\textit{i}.\textit{e}.}
\newcommand{\eg}{\textit{e}.\textit{g}.}

\acmSubmissionID{2785}

\begin{document}

\title{Rethinking the Reference-based Distinctive Image Captioning}

\author{Yangjun Mao$^{*}$}
\affiliation{%
    \authornote{Yangjun Mao and Long Chen are co-first authors with equal contributions to this work. Our code is available on \href{https://github.com/maoyj1998/TransDIC}{https://github.com/maoyj1998/TransDIC}.}
    \institution{Zhejiang University}
    \city{}
    \country{}
 }
\email{maoyj0119@zju.edu.cn}

\author{Long Chen$^{*}$}
\affiliation{%
    \institution{Columbia University}
    \city{}
    \country{}
  }
\email{zjuchenlong@gmail.com}

\author{Zhihong Jiang}
\affiliation{
    \institution{Zhejiang University}
    \city{}
    \country{}
}
\email{zju_jiangzhihong@zju.edu.cn}

\author{Dong Zhang}
\affiliation{
    \institution{Hong Kong University of Science and Technology, dongz@ust.hk}
    \city{}
    \country{}
}

\author{Zhimeng Zhang, Jian Shao}
\affiliation{
    \institution{Zhejiang University}
    \city{}
    \country{}
}
\email{zhimeng@zju.edu.cn, jshao@zju.edu.cn}


\author{Jun Xiao$^{\dag}$}
\affiliation{
    \authornote{Jun Xiao is the corresponding author.}
    \institution{Zhejiang University}
    \city{}
    \country{}
}
\email{junx@zju.edu.cn}

\renewcommand{\shortauthors}{Yangjun Mao, et al.} 


\begin{abstract}
Distinctive Image Captioning (DIC) --- generating distinctive captions that describe the unique details of a target image --- has received considerable attention over the last few years. A recent DIC work proposes to generate distinctive captions by comparing the target image with a set of semantic-similar reference images, \ie, reference-based DIC (Ref-DIC). It aims to make the generated captions can tell apart the target and reference images. Unfortunately, reference images used by existing Ref-DIC works are easy to distinguish: \emph{these reference images only resemble the target image at scene-level and have few common objects, such that a Ref-DIC model can trivially generate distinctive captions even without considering the reference images.} For example, if the target image contains objects ``\texttt{towel}'' and ``\texttt{toilet}'' while all reference images are without them, then a simple caption ``\texttt{A bathroom with a towel and a toilet}'' is distinctive enough to tell apart target and reference images. To ensure Ref-DIC models really perceive the unique objects (or attributes) in target images, we first propose two new Ref-DIC benchmarks. Specifically, we design a two-stage matching mechanism, which strictly controls the similarity between the target and reference images at object-/attribute- level (vs. scene-level). Secondly, to generate distinctive captions, we develop a strong Transformer-based Ref-DIC baseline, dubbed as \textbf{TransDIC}. It not only extracts visual features from the target image, but also encodes the differences between objects in the target and reference images. Finally, for more trustworthy benchmarking, we propose a new evaluation metric named \textbf{DisCIDEr} for Ref-DIC, which evaluates both the accuracy and distinctiveness of the generated captions. 
Experimental results demonstrate that our TransDIC can generate distinctive captions. Besides, it outperforms several state-of-the-art models on the two new benchmarks over different metrics. 


\end{abstract}





\maketitle

\begin{figure}[t]
\centering
\includegraphics[width=\linewidth]{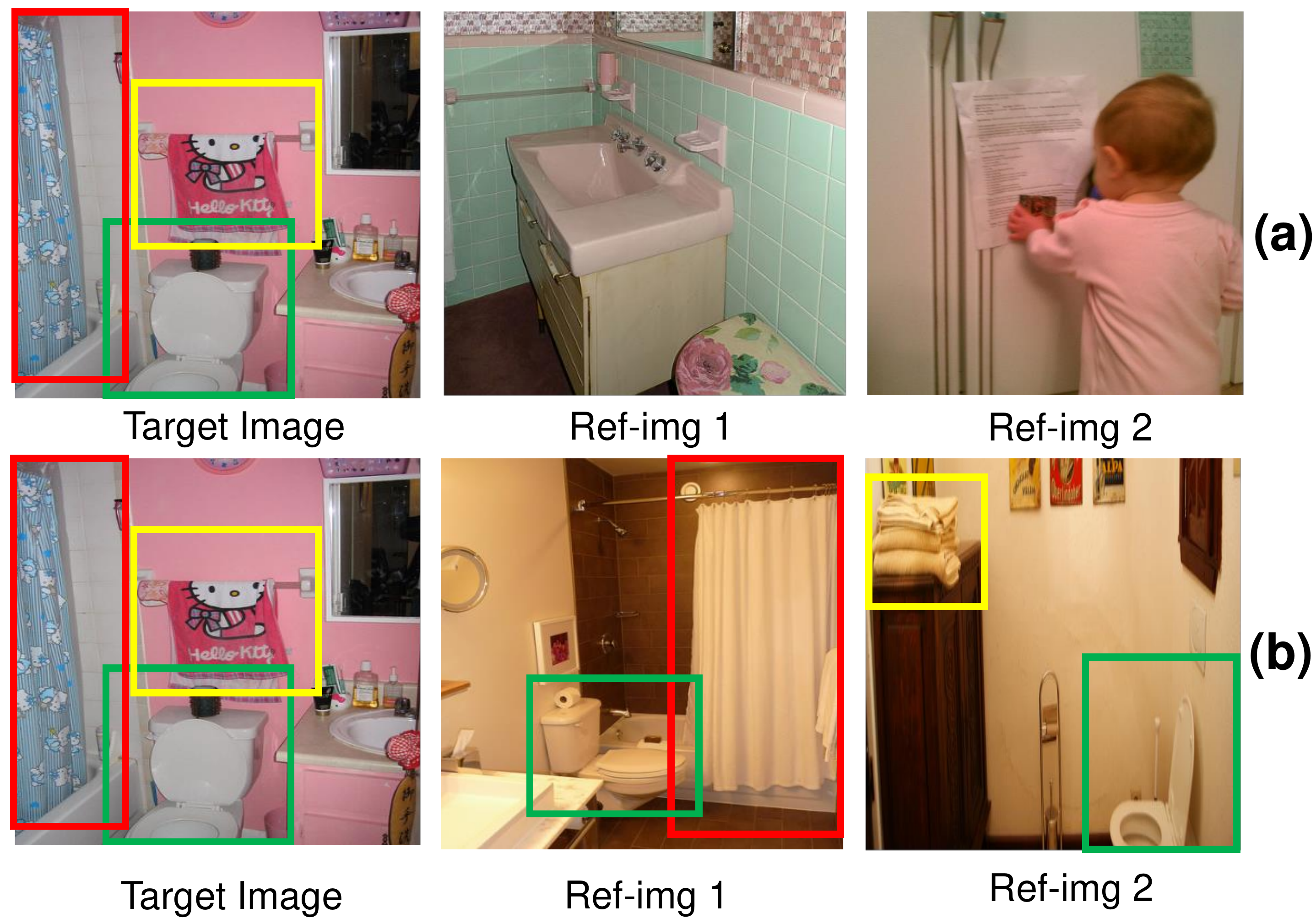}
\caption{(a): An example of constructed reference image group used in existing Ref-DIC work~\cite{wang2021group}. (b): Selected reference images for the same target image using our two-stage matching mechanism. We use same colors to denote the same object categories in the different images (\eg, ``\texttt{towel}'' is with \textcolor{myyellow}{yellow} box, ``\texttt{shower curtain}'' is with \textcolor{myred}{red} box, and ``\texttt{toilet}'' is with \textcolor{mygreen}{green} box). We only show two reference images here.}
\label{fig_1_dataset_example}
\end{figure}

\section{Introduction}

Image captioning, \ie, generating natural language descriptions to summarize the salient contents of a target image, has drawn much attention from the multimedia community. It has great impacts on many downstream applications, such as helping blind people and developing navigation systems. However, as revealed in~\cite{dai2017towards, dai2017contrastive}, conventional image captioning models tend to generate over-generic captions or even identical captions when input images are similar. Obviously, these generic captions neglect the unique details of the target image. Recent captioning works~\cite{dai2017contrastive, luo2018discriminability, liu2018show, wang2020compare} begin to make generated captions more distinctive and ask these captions describe more unique details of each target image, called Distinctive Image Captioning (DIC).

Currently, mainstream DIC works follow the same setting as plain image captioning: using one single image as input, and generating distinctive captions for each image, dubbed as Single-image DIC (\textbf{Single-DIC}). In this setting, they tend to generate a totally distinctive caption. By ``totally'', we mean the generated caption is asked to distinguish its corresponding image from all images in the dataset, \ie, dataset-level distinctiveness. To this end, they always resort to reinforcement learning and develop different distinctive rewards~\cite{luo2018discriminability,liu2018show}. However, this Single-DIC setting has two inherent issues: 1) It is difficult (or impossible) to generate a totally distinctive caption for the target image unless we describe all the details in the image. 2) Even for our humans, we still need some reference images when generating distinctive captions. For example in Figure~\ref{fig_1_dataset_example} (b), without any reference images, people won't know what should be emphasized in the \emph{target image}, and may simply predict ``\texttt{A bathroom with a towel}'' for the image. In contrast,  they will focus on the unique colors of ``\texttt{towel}'' and ``\texttt{shower curtain}'', and predict ``\texttt{A bathroom with a pink towel and a blue shower curtain}'' when they use the ``\texttt{white shower curtain}'' in \emph{ref-img1} and ``\texttt{yellow towel}'' in \emph{ref-img2} as references.

For human-like distinctive captioning, a recent work proposes to study the DIC task based on a group of semantic-similar reference images, dubbed Reference-based DIC (\textbf{Ref-DIC}).
Different from Single-DIC, they use the target image and all reference images as input and these reference images will inform DIC models which parts of the target image should be emphasized. Compared to Single-DIC, the generated captions are only asked to distinguish the target image from the group of reference images, \ie, group-level distinctiveness. Unfortunately, the reference images used in existing Ref-DIC works~\cite{wang2021group} can be trivially distinguished: \emph{these reference images only resemble the target image at the scene-level and have few common objects, thus Ref-DIC models can simply generate distinctive captions even without considering the reference images.} For example in Figure~\ref{fig_1_dataset_example} (a), \emph{target} and \emph{reference images} have no object in common (\eg, ``\texttt{towel}", ``\texttt{shower curtain}", or ``\texttt{toilet}"), each object in \emph{target image} is unique, such that the Ref-DIC model can trivially generate ``\texttt{a bathroom with a towel}'' to tell the target and reference images apart.

As mentioned above, reference images are crucial when defining the unique details in the target image. In this paper, we propose two new benchmarks for the Ref-DIC task: \textbf{COCO-DIC} and \textbf{Flickr30K-DIC}.
To strictly control the unique details between target and reference images, we propose a two-stage matching mechanism, which can measure image similarity at the object-/attribute- level (vs. scene-level in~\cite{wang2021group}), and deliberately make target and reference images have some common objects. Under this mechanism, Ref-DIC models can learn to focus on the unique attributes and objects in target image. As the example in Figure~\ref{fig_1_dataset_example} (b), compared to \emph{ref-img1}, \emph{target image} has the unique attribute ``\texttt{blue}'' of ``\texttt{shower curtain}'' and the unique object ``\texttt{towel}''.

To achieve group-level distinctiveness, we propose to emphasize both unique attributes and objects in target image. Thus, we also propose a new Transformer-based captioning model named \textbf{\model}, which directly gives each region in target image (target regions) some region references when generating captions. Specifically, we firstly find similar regions from reference images (reference regions) for each target region. Then, we send target region and its corresponding reference regions into the \emph{\flows}~Module, which consists of a Target flow and a Target-Reference flow. The Target flow aims to encode target image features through self-attention blocks~\cite{vaswani2017attention}. And the Target-Reference flow enables cross-image interactions between target and reference images through a multi-layer co-attention. Different from the existing Ref-DIC work~\cite{wang2021group} which proposes an attention module to focus on unique objects in target image, our TransDIC directly enables the feature interactions between target and reference images. 

Furthermore, to fully take advantage of ground-truth captions of reference images, we propose a new CIDEr-based~\cite{vedantam2015cider} metric termed as \textbf{DisCIDEr}. According to our definition of group-level distinctiveness, we believe frequently-used n-grams in ground-truth captions of reference images should be given less weight at evaluation time. The metric can not only directly evaluate the distinctiveness, but also preserve the accuracy advantage of CIDEr. Extensive experimental results on multiple Ref-DIC benchmarks (COCO-DIC and Flickr30K-DIC) have demonstrated the effectiveness of our proposed \model.

In summary, we make three contributions in this paper:

\begin{enumerate}[topsep=0pt,leftmargin=2em]
	\item We proposed two new benchmarks for Ref-DIC, constructed groups in those benchmarks can inform the model which parts in target image should be emphasized.
	\item We developed a new model \model~for Ref-DIC, which achieves better performance than the state-of-the-art Ref-DIC models in terms of both accuracy and distinctiveness.
	\item  We proposed a new metric named DisCIDEr which considers both the distinctiveness and accuracy of the generated caption.
\end{enumerate}

\section{Related Work}
\noindent\textbf{Image Captioning.}  Most modern image captioning models typically employ an encoder-decoder framework for caption generation~\cite{vinyals2015show,karpathy2015deep,johnson2016densecap,liu2021region,wang2021high}. Within this framework, many efforts have been made to improve the architecture, including attention mechanisms~\cite{xu2015show,chen2017sca,anderson2018bottom,huang2019attention,chen2021human}, graph convolution networks~\cite{yao2018exploring,yang2019auto,chen2020say}, and transformer-based models~\cite{li2019entangled,2020Meshed}. Meanwhile, another series of works explore different training objectives at the training stage. For example, Dai~\etal~\cite{dai2017towards} and Shetty~\etal~\cite{shetty2017speaking} leverage Generative Adversarial Network (GAN) to the improve the diversity of generated captions. Some recent captioning works apply Reinforcement Learning (RL) to captioning and achieve great success~\cite{ranzato2015sequence,rennie2017self,liu2017improved,xu2019multi}. These models directly optimize  non-differentiable 
evaluation metrics (\eg, BLEU~\cite{papineni2002bleu}, CIDEr~\cite{vedantam2015cider}), which boost the caption generation procedure at the sentence-level.

\begin{figure*}[t]
\centering
\includegraphics[width=\linewidth]{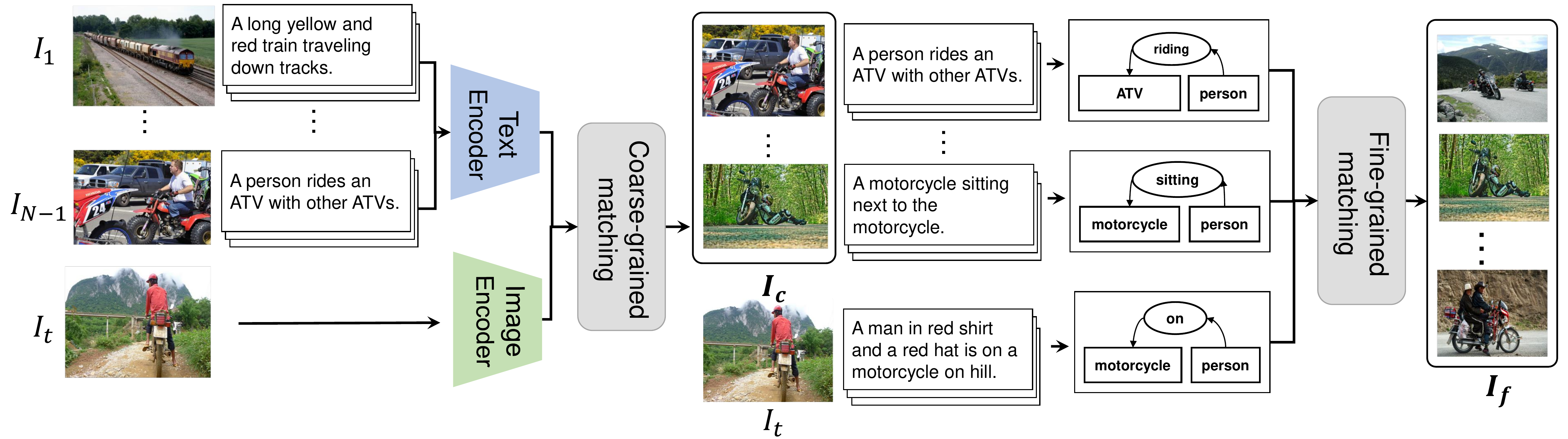}
\caption{The pipeline of our two-stage matching procedure. In the first stage, we calculate image-text similarity scores between target image $I_t$ and the captions of all other images in $\mathcal{D}$ through CLIP~\cite{radford2021learning} and construct a coarse-grained group $\mathcal{I}_c$ for target image $I_t$. In the second stage, we leverage scene graphs to calculate the object and attribute overlaps between images in $\mathcal{I}_c$ and $I_t$. We rearrange $\mathcal{I}_c$ according to their similarity scores with $I_t$, and finally get the fine-grained reference image group $\mathcal{I}_f$.}
\label{fig_2_group_construction}
\end{figure*}

\noindent\textbf{Distinctive Image Captioning (DIC).} Compared with conventional image captioning, DIC is a more challenging task, which tends to generate more informative and descriptive captions.
According to the stage they take effect, existing solutions can be coarsely divided into two categories: \emph{Inference-based} and \emph{Training-based} methods. Inference-based models mainly modify the caption decoding procedure at inference time and thus can be applied to any captioning architecture~\cite{vedantam2017context,wang2020towards}. In contrast, training-based methods, resort to different training objects~\cite{luo2018discriminability,liu2018show} or the progressive training procedure~\cite{liu2019generating}. Recently, some works begin to study the DIC task based on semantic-similar reference images. Specifically, Wang~\etal~\cite{wang2020compare} propose to assign higher weights to distinctive ground-truth captions at the training stage, Wang~\etal~\cite{wang2021group} use multiple images as input to emphasize distinctive objects. In this paper, we propose a co-attention based model to directly enable the feature interactions between target and reference images.

To evaluate the distinctiveness of generated captions in DIC, several new evaluation metrics are developed.  SPICE-U~\cite{wang2020towards} is designed for Single-DIC. CIDErBtw~\cite{wang2020compare} measures the distinctiveness of caption at sentence-level similarity.  DisWordRate~\cite{wang2021group} directly evaluates the occurrences of distinctive words. In this paper, we develop a new metric named DisCIDEr for Ref-DIC. Compared to existing metrics, our metric fully explore the distinctiveness of each individual n-gram in ground-truth captions of target image.

\noindent\textbf{Multi-input Image Captioning.} Several captioning settings need multiple images as input. According to the number of input images, they can be divided into two categories: \emph{Two-image based} and \emph{Group-based} captioning.
Two-image based captioning tends to describe the common~\cite{suhr2018corpus} or different~\cite{tan2019expressing,park2019robust,yan2021l2c,qiu2021describing} parts between the two images. Thus, the two images in their settings always have strong correlations. For example, the change captioning task takes before and after images as input and describes the changes between them.
In contrast, group-based captioning typically uses a group of images as references to investigate certain properties of the target image. For example, Chen~\etal~\cite{chen2018groupcap} firstly model the relevance and diversity between target and reference images and aim to generate diverse captions for the target image. Li~\etal~\cite{li2020context} tend to describe a group of target images using another group of 
semantically similar images as references.

\section{Proposed Benchmarks}
In this section, we firstly formally define the Ref-DIC task. Then we describe our solution for Ref-DIC benchmarks construction. Finally, we provide details of our proposed COCO-DIC and Flickr30K-DIC benchmarks for Ref-DIC.

\subsection{Task Definition: Reference-based DIC}
Given a \textbf{target image} $I_t$ and a group of $K$ \textbf{reference images} $\mathcal{I}_r=\{I_i\}_{i=1}^K$ which are semantic-similar to $I_t$, Ref-DIC models aim to generate a natural language sentence $S=\{w_1,w_2,\ldots,w_T\}$. The generated sentence $S$ should not only correctly describe the target image $I_t$, but also contain sufficient details about $I_t$, so it can tell apart target and  reference images. For example in Figure~\ref{fig_1_dataset_example} (b), given the target and reference images, Ref-DIC models aim to generate a distinctive caption ``\texttt{a bathroom with a pink towel, a blue shower curtain and a toilet}''. The detail ``\texttt{pink towel}'' is helpful to distinguish target image from \emph{ref-img2}~because the ``\texttt{towel}'' in \emph{ref-img2} is ``\texttt{white}''. On the contrary, predicting  ``\texttt{a bathroom with a towel and a toilet}'' fails to meet the requirements because it is suitable for both target and reference images.

\subsection{Ref-DIC Benchmarks Construction} \label{sec:sim_group}
Given a conventional image captioning dataset $\mathcal{D}$, suppose it contains $N$ images and each one has $M$ corresponding ground-truth captions. We build new Ref-DIC benchmarks based on $D$, by coupling each image (target image) with several semantic-similar reference images.
Specifically, each image in $\mathcal{D}$ will be regarded as a target image $I_t$, and all remaining $N-1$ images are termed as its \textbf{candidate reference images}. For each target image $I_t$, our goal is to retrieve $K$ reference images from its candidate reference images to construct the reference image group $\mathcal{I}_r$.

To achieve group-level distinctiveness, $I_t$ and retrieved $\mathcal{I}_r$ should have some common objects, such that $\mathcal{I}_r$ will inform the model to focus on the unique details in $I_t$. To this end, we design a two-stage matching mechanism. In the first stage, we construct a \textbf{coarse-grained group} $\mathcal{I}_c$ based on the image-text similarity score for each target image. Then in the second stage, we investigate fine-grained details of $I_t$ and $\mathcal{I}_c$, and construct a \textbf{fine-grained group} $\mathcal{I}_f$ based on $\mathcal{I}_c$. Finally, we select $K$ images out of $\mathcal{I}_f$ to construct the $\mathcal{I}_r$. We detailed introduce our two-stage matching pipeline below.

\subsubsection{Coarse-grained Group Construction} \label{sec:candidate group}
Following~\cite{wang2021group}, we use an image-text retrieval model to calculate similarity scores between images and texts. Specifically, as shown in Figure~\ref{fig_2_group_construction} (left), we use a pre-trained  CLIP~\cite{radford2021learning} to firstly extract the visual feature of target image $I_t$ and text features of ground-truth captions from candidate reference images. Then, we perform the cosine similarity between visual feature and all text features to get $(N-1)\times M$ scores. Finally, we select $||\mathcal{I}_c||$ captions with the highest scores, and their corresponding images are used as the coarse group $\mathcal{I}_c$ for $I_t$.

The CLIP-based matching mechanism can effectively filter out some obviously unrelated candidate images. However, since it encodes texts at the sentence level, several fine-grained details may be neglected when computing similarity. For example in Figure~\ref{fig_2_group_construction}, ground-truth captions of candidate image $I_1$ contain the  ``\texttt{train}'', thus $I_1$ will be removed due to a low scene-level similarity score to $I_t$. Meanwhile, $I_{N-1}$ is considered similar to $I_t$ by the CLIP because both of them describe the scene of ``\texttt{someone is riding a vehicle}''. However, they resemble each other only at the sense-level and do not contain any common objects (\eg, ``\texttt{motorcycle}'' in $I_{t}$ and ``\texttt{ATV}'' in $I_{N-1}$ are totally different objects).

\begin{figure}[t]
\centering
\includegraphics[width=\linewidth]{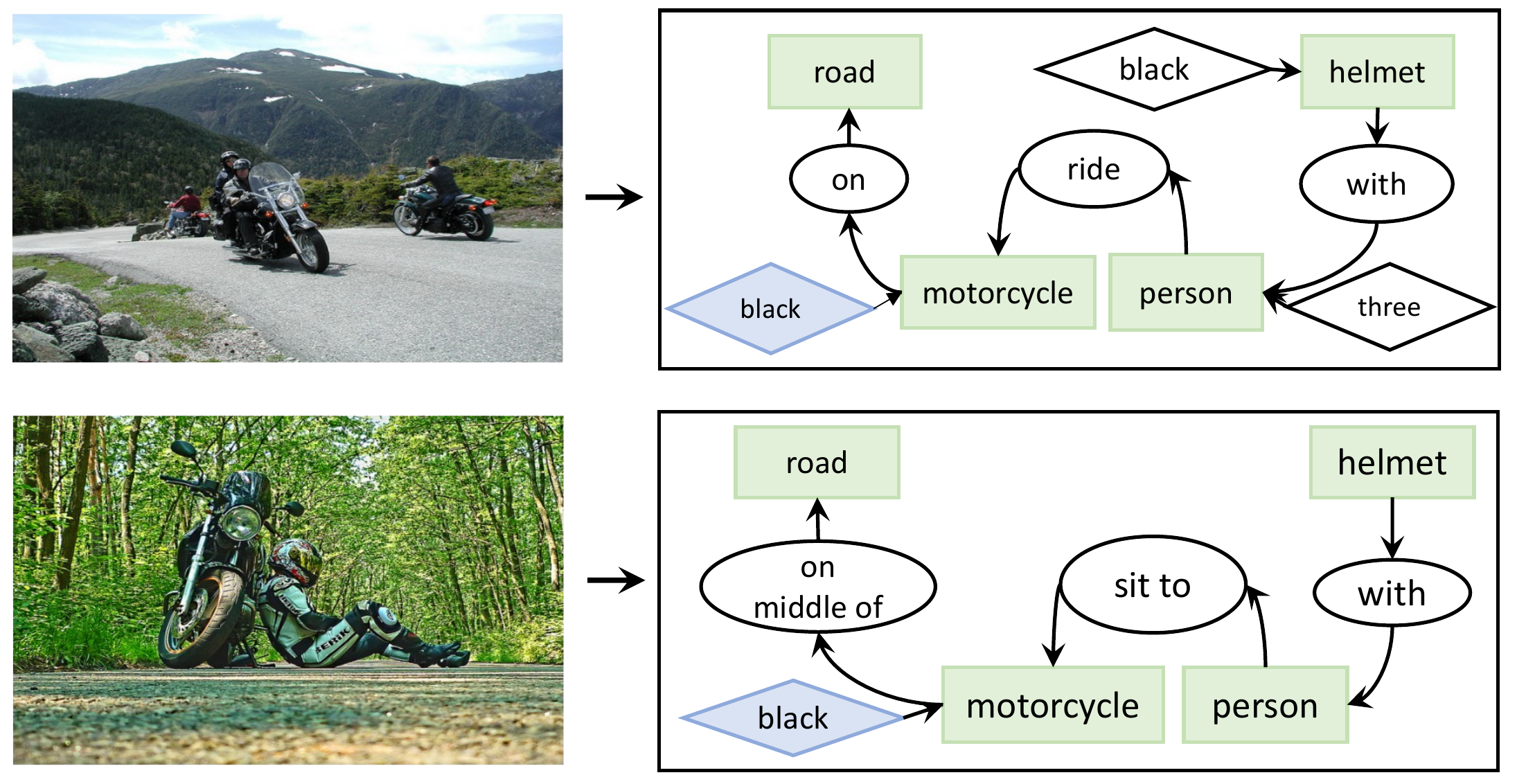}
\caption{An example of parsed scene graphs for the ground-truth captions of two images. Two graphs have four object overlaps: ``\texttt{helmet}'', ``\texttt{people}'', ``\texttt{motorcycle}'', and ``\texttt{road}'' (\textcolor{mygreen}{green}), and one attribute overlap ``\texttt{black}'' (\textcolor{myblue}{blue}).}
\label{fig_3_fine_grained_matching}
\end{figure}

\addtolength{\tabcolsep}{-1pt}
\begin{table}[t]
  \centering

    \caption{Statistical summary of the COCO-DIC, Flickr30K-DIC, and existing Ref-DIC benchmark~\cite{wang2021group}. ``\#overlaps'' denotes the number of object/attribute overlap in each dataset.}
    \begin{tabular}{l|cccc|ccc}
    \hline
    \multirow{2}[2]{*}{Datasets} & \multicolumn{4}{c|}{images} & \multicolumn{3}{c}{\#overlaps in a group}\\
    \multicolumn{1}{c|}{} & 
    \multicolumn{1}{c}{Train} & \multicolumn{1}{c}{Val} & \multicolumn{2}{c|}{Test} & \multicolumn{1}{c}{Train} & \multicolumn{1}{c}{Val} & \multicolumn{1}{c}{Test} \\
    \hline
    Wang~\etal~\cite{wang2021group} & 133,980 & 5,562  & \multicolumn{2}{c|}{5,538} & 3.8   & 3.7   & 3.7 \\
    COCO-DIC & 123,287 & 5,000  & \multicolumn{2}{c|}{5,000} & 5.0     & 4.9   & 4.9 \\
    Flickr30K-DIC & 29,000 & 1,014  & \multicolumn{2}{c|}{1,000} & 5.3   & 5.3   & 5.3 \\
    \hline
    \end{tabular}%
 
   \label{table:dataset_statics}%
\end{table}%
\addtolength{\tabcolsep}{1pt}

\subsubsection{Fine-grained Group Construction}
To overcome the shortcoming of the coarse-grained group, we propose a fine-grained matching mechanism that directly uses object and attribute overlaps between two images as the similarity measurement. Firstly, for $I_t$ or any image in $\mathcal{I}_c$, we parse all its ground-truth captions into one scene graph. Then, we extract objects and attributes from scene graphs~\cite{chen2019counterfactual,xu2020scene,li2022devil,liu2021toward} of two images to calculate overlaps. Specifically, objects from the two graphs will be compared according to their categories. However, two attributes should firstly correspond to the same objects and then compare to each other. For example in Figure~\ref{fig_3_fine_grained_matching}, when calculating object overlaps, ``\texttt{black helmet}'' (top) and ``\texttt{helmet}'' (bottom) from two graphs denote one-time object overlap (\eg, common object ``\texttt{helmet}''). As for attribute overlaps, two graphs have  both ``\texttt{black motorcycle}'' in common and denote one-time attribute overlap\footnote{Note that the ``\texttt{black helmet}'' and  ``\texttt{black motorcycle}'' contain no attribute overlap because the attribute ``\texttt{black}'' belongs to different objects.}. In this paper, we take the sum over object and attribute overlaps as the final similarity score for two images. And we sort all images in $\mathcal{I}_c$ according to their similarity scores to $I_t$ to construct the fine-grained group $\mathcal{I}_f$. It is worth noting that we select $K$ images but not the top-K from $\mathcal{I}_f$ to construct $I_r$. The reason for this choice is that we believe the most similar images from $\mathcal{I}_f$ may contain the identical objects and attributes as $I_t$, thus they won't help to emphasize any unique details in $I_t$. More detailed discussion about the top-K selection are left in Table~\ref{table:abal_group}.

\subsection{Benchmarks: COCO-DIC \& Flickr30K-DIC}
We apply our matching mechanism to widely-used captioning benchmarks MS-COCO~\cite{2015Microsoft} and Flickr30K~\cite{2016Flickr30k} to construct \textbf{COCO-DIC} and \textbf{Flickr30K-DIC}, respectively. Some basic statistics about our proposed benchmarks are reported in Table~\ref{table:dataset_statics}. Different from the construction procedure proposed in~\cite{wang2021group}, they avoid image reuse (or overlap) among different constructed groups. Thus, some images which are not similar enough may be forced to construct a group. In contrast, we find $K$ reference images independently to ensure the similarity within a group.

\section{Proposed Approach}

\subsection{Preliminaries}
\subsubsection{Transformer-based Image Captioning}
Transformer~\cite{vaswani2017attention} follows the standard \textbf{Encoder-Decoder} architecture. It employs the self-attention mechanism to explore the internal correlation within the sequential data, which has been widely adopted by numerous image captioning models~\cite{li2019entangled,2020Meshed}. 
For a given image $I$, they use proposal features~\cite{anderson2018bottom} extracted by an object detector as input: $X=\{x_i\}_{i=1}^N$ where $x_i \in \mathbb{R}^d$ is the feature vector for $i$-th proposal in $I$ and $N$ is the number of proposals. They employ multiple self-attention layers as \textbf{Encoder}, and the outputs of the $l$-th layer is calculated as follows:
\begin{align} \label{self-attn layers}
		H_{l-1} & =\textbf{LN}\left( O_{l-1}+\textbf{MH}( O_{l-1}, O_{l-1}, O_{l-1}) \right), \\
		O_{l}  & = \textbf{LN}\left( H_{l-1} + \textbf{FFN}(H_{l-1}) \right),
\end{align}
\noindent where $O_0$ refers to input proposal features $X$, and $O_{l-1}$ is outputs of the $(l-1)$-th layer. $\textbf{LN}(\cdot)$ denotes the layer normalization~\cite{2016Layer}, $\textbf{FFN}(\cdot)$ denotes the feed forward network, and $\textbf{MH}(\cdot)$ denotes the multi-head attention~\cite{vaswani2017attention}. Encoded visual features are fed into the \textbf{Decoder} for caption generation. It generates a word probability distribution $P_t = P(w_t|w_{1:t-1},I)$ at each time step t conditioning on the previously generated words $\{w_1,\ldots,w_{t-1}\}$ and image $I$.

\subsubsection{Model Optimization}
Mainstream captioning works typically resort to a two-stage procedure for model optimization~\cite{rennie2017self,2020Meshed}. Given an image $I$ and its ground-truth captions $C = \{c_i\}_{i=1}^M$. They firstly apply a cross-entropy loss (XE) to pre-train the model and then employ reinforcement learning (RL) to finetune sequence generation. 

When training with XE, for a ground-truth caption $c_i = \{w_t\}_{t=1}^T$, they ask the model to minimize the following cross-entropy loss:
\begin{equation}
    L_{xe} = - \sum_{t=1}^T\log P(w_t|w_{1:t-1},I),
\end{equation}
where $P(w_t|w_{1:t-1},I)$ denotes the predicted probability of word $w_t$.

When training with reinforcement learning, they firstly generate top-n captions $\hat{C} = \{\hat{c}_i\}_{i=1}^n$ through beam search, and then optimize the following RL loss~\cite{2020Meshed}\footnote{Note that there are several RL variants for caption generation, we only demonstrate one typical policy gradient solution here.}:
\begin{equation}
    L_{rl} = - \frac{1}{n} \sum_{i=1}^n ((r(\hat{c}_i, C) - b) \log p(\hat{c}_i)),
\end{equation}
where r($\cdot,\cdot$) is the reward function computed between $\hat{c}_i$ and $C$, and $b = (\sum_{i=1}^n r(\hat{c}_i,C)) / n$ is the baseline, calculated as the mean of the rewards obtained by the generated captions.

\begin{figure*}[t]
\centering
\includegraphics[width=0.97\linewidth]{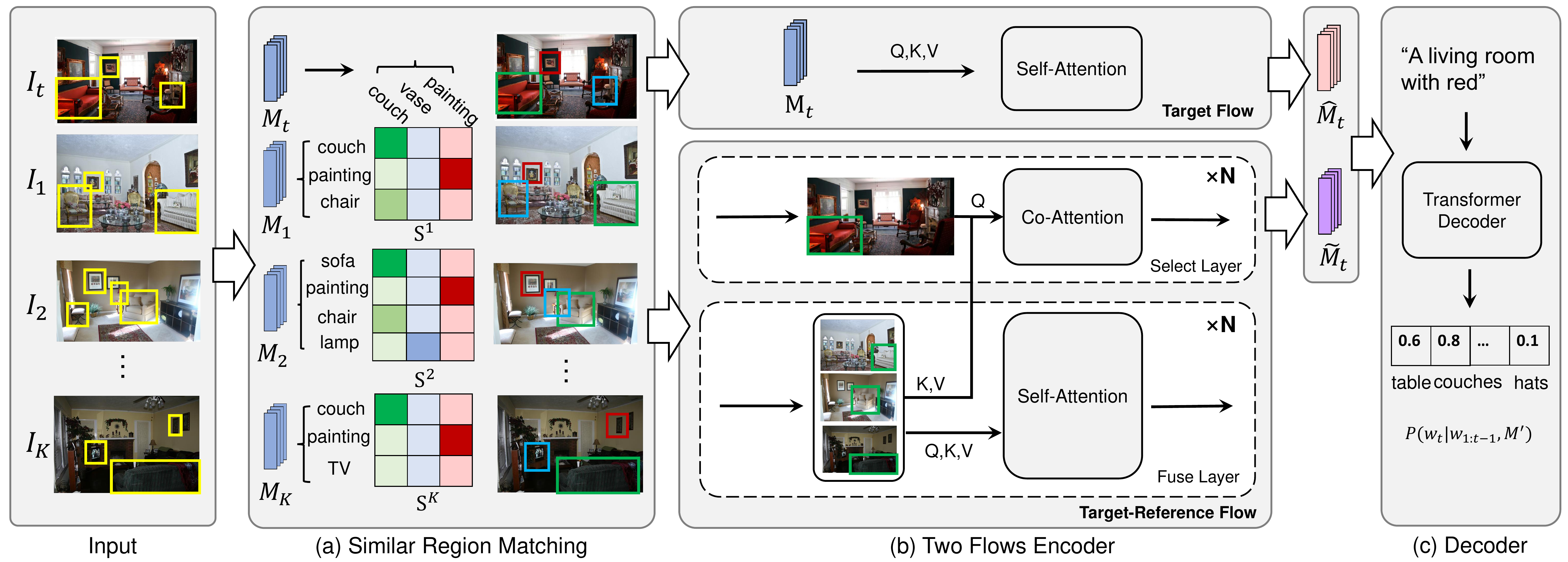}
\caption{Overview of our proposed \model~model. It consists of three parts: (a) A Similar Region Matching module that couples each target region with reference regions. (b) A \flows~module that encodes both target and reference images. (c) A plain captioning decoder. In module (a), we use the same colors to denote the same objects. We firstly send target and reference images into the similar region matching module to construct Target-Reference tuples, \eg, all regions marked with green boxes contain the object ``\texttt{couch}'', thus form a Target-Reference tuple. Then we send constructed tuples into the \flows~for target image and cross-image features extraction. Finally, both kinds of features are sent into the decoder for caption generation.}
\label{fig_4_model_overview}
\end{figure*}

\subsection{\model: Transformer-based Ref-DIC}
Given a target image $I_t$, we term all regions (or proposals) in it as \textbf{target regions} $R^t=\{r_n^t\}_{n=1}^N$. Our proposed model tends to give each target region $r_n^t$ some region references when generating captions. To this end, we firstly couple each $r_n^t$ with semantic-similar regions from $\mathcal{I}_r$ as the reference, \ie, \textbf{reference regions}. Then we send each target region and its reference regions into the model for distinctive caption generation.


Specifically, our \model~consists of three components: 
1) Similar Region Matching module in Section~\ref{sec: sim_box_matching}. 2) \flows~module. The module contains two parallel data flows to extract both target image and cross-image features in Section~\ref{sec:select-fuse}. 
3) A plain Transformer-based Captioning Decoder for caption generation. An overview of our model is shown in Figure~\ref{fig_4_model_overview}.

\subsubsection{Similar Region Matching} \label{sec: sim_box_matching}
For each target region, we retrieve regions from reference images with the highest similarity scores to it as its reference regions. Given a target image $I_t$ and its corresponding $\mathcal{I}_r$, their region features are firstly projected into the memory space through an MLP layer. We
denote memory features for $I_t$ and the K images in $\mathcal{I}_r$ as $M_t =\{m^t_j\}_{j=1}^N$ and $M_{k}=\{m^k_i\}_{i=1}^N,k=\{1\dots K\}$ , respectively.

Then, we calculate the cosine similarity scores between features in $M_t$ and $M_k$, \ie, 
\begin{equation}
    S^k_{ij} = cos(m^k_{i},m^t_j),
\end{equation}
where $m^t_j$ represents the $j$-th region in $I_t$, and $m^k_{i}$ represents the $i$-th region in reference image $I_k$. We apply max operation to get the most similar region for $r_j^t$ according to the calculated $S^k$:
\begin{equation}
    \hat{r}^k_j = \mathop{\arg\max}_i(\{S^k_{ij}\}_{i=1}^N),
\end{equation} 
where $\hat{r}^k_j$ denotes the most similar region from image $I_k$ for $r_j^t$, \ie, reference region.
As an example in Figure~\ref{fig_4_model_overview} (a), reference image $I_K$ contains regions ``\texttt{couch}'', ``\texttt{painting}'' and ``\texttt{TV}''. For the region ``\texttt{couch}'' in $I_t$, we can learn from the similarity matrix $S^K$: the region is similar to the ``\texttt{couch}'' in $I_K$ (deep green) while is different from the ``\texttt{TV}'' or ``\texttt{painting}'' (light green) in $I_K$. Max operation is then token along each column of $S^K$, and ``\texttt{couch}'' in $I_k$ is selected as the reference region for ``\texttt{couch}'' in $I_t$.

Finally, for each target region $r_n^t$, we gather $K$ reference regions, one for each, from $K$ reference images. We put these $K+1$ regions together and term them as a \textbf{Target-Reference region tuple}:
\begin{equation}
    T_n = \{r_n^t,\hat{r}_n^1,\hat{r}_n^2,\dots,\hat{r}_n^K \}. \quad n= \{1, \dots ,N \}
\end{equation}
For example in Figure~\ref{fig_4_model_overview} (a), all regions marked with green boxes form a Target-Reference tuple (for ``\texttt{couch}'').

\subsubsection{\flows} \label{sec:select-fuse}
Our proposed  module takes $M_t$ and $M_k$ as input, and extracts target image features and cross-image features through the \textbf{Target flow} and the \textbf{Target-Reference flow},  respectively. An overview of the module is shown in~\ref{fig_4_model_overview} (b).


\textbf{Target flow.} The flow enables the region interactions within the target image $I_t$. Same as the standard transformer-based captioning model, it sends memory features $M_t$ into multiple self-attention layers, and finally outputs encoded features $\hat{M}_t=\{\hat{m}^t_i\}_{i=1}^N$ for $I_t$.

\textbf{Target-Reference flow.} This data flow consists of \textbf{select layers} and \textbf{fuse layers}. Given a Target-Reference region tuple $T_n$, we denote the memory features for target region and reference regions in $T_n$ as $\bar{m}_n^t$ and $\bar{M}_n = \{\bar{m}_n^i\}_{i=1}^K$, respectively. The flow takes in those two kinds of features and generates cross-image features through \textbf{select} and \textbf{fuse} layers:

\textbf{\emph{Fuse layer}}. The goal of the fuse layer is to enable the interactions among memory features within $\bar{M}_n$. We stack multiple fuse layers, and the $l$-th fuse layer is calculated as follows:
\begin{equation}
    U_{l} = \textbf{MH}(U_{l-1},U_{l-1} ,U_{l-1}),
\end{equation}
where $U_{0}$ refers to $\bar{M}_n$ and $U_{l-1}$ is the outputs of the $(l-1)$-th fuse layer. 
Because all the features in $\bar{M}_n$ are semantic-similar, the model can learn to capture the primary concepts they are describing. 

\textbf{\emph{Select layer}}. The select layer builds on the co-attention mechanism. We set features of target region as query, features of reference regions as key and value in multi-head attention. Multiple co-attention layers are stacked, the $l$-th select layer is computed as:
\begin{equation}
    	V_{l} = \textbf{MH}(V_{l-1},U_{l-1} ,U_{l-1}),
\end{equation}
where $V_{0}$ refers to $\bar{m}_n^t$, $V_{l-1}$ and $U_{l-1}$ are the outputs of the $(l-1)$-th select and fuse layer, respectively. By the residual connection in self-attention blocks, feature $\bar{m}_n^t$ will gradually select useful information from reference images while preserving the original information from $I_t$.

As an example shown in Figure~\ref{fig_4_model_overview} (b), our model can learn to focus on the unique attributes and objects in $I_t$. For unique attributes, we send all reference regions of ``\texttt{couch}'' (green boxes) into fuse layers, the model will be informed they are describing the concept ``\texttt{couch}''. Since target region also describes ``\texttt{couch}'', the select layer learns to focus on the unique color ``\texttt{red}'' of the ``\texttt{couch}'' in $I_t$. When predicting unique objects, for ``\texttt{vase}'' region in $I_t$, because all selected reference regions for it (blue boxes) do not contain the same concept, the select layer learns that ``\texttt{vase}'' is a unique object in $I_t$.

We use the outputs of the last select layer as final refined target feature $\tilde{m}_n^t$ for region $r_n^t$ in $I_t$. For N Target-Reference tuples in $I_t$, we can get $\tilde{M}_t =  \{\tilde{m}_i^t\}_{i=1}^N$ as the outputs of Target-Reference flow. Finally, we concatenate the outputs of Target flow and Target-Reference flow as the final outputs of the \flows:
\begin{equation}
    M_t' = [\hat{M}_t;\tilde{M}_t],
\end{equation}
where $[\cdot;\cdot]$ denotes concatenation operation, and $M_t'$ will be sent into decoder for caption generation. 

\begin{figure}[t]
\centering
\includegraphics[width=\linewidth]{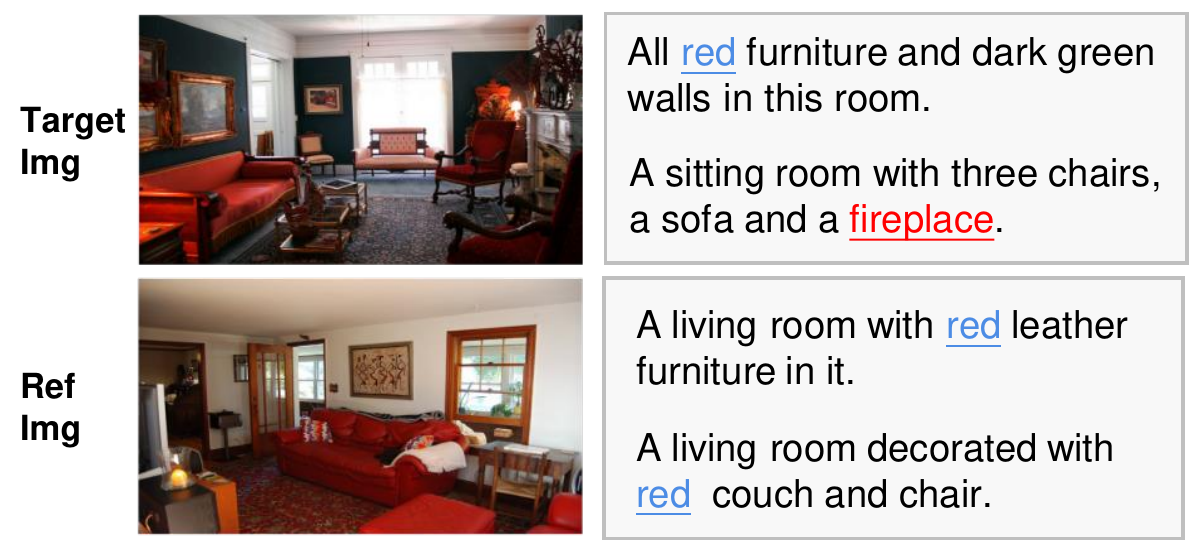}
\caption{An example of the intuition behind DisCIDEr. N-gram ``\texttt{red}'' appears in both target and reference images, thus should be given less attention (\textcolor{myblue}{blue}). In contrast, we should pay more attention to ``\texttt{fireplace}'', because it  appears only in target image (\textcolor{red}{red}).}
\label{fig_5_discider_example}
\end{figure}

\section{Experiments}
In this section, we describe the datasets used for experiments and introduce a new distinctiveness-based evaluation metric \textbf{DisCIDEr}. We conduct extensive experiments and ablation studies to reveal the superiority of our proposed model, as well as our proposed benchmarks for Ref-DIC.

\subsection{Datasets}
We developed the \textbf{COCO-DIC} and \textbf{Flickr30K-DIC} based on the MS-COCO~\cite{2015Microsoft} and Flickr30K~\cite{2016Flickr30k}. They contain 123287 and 31014 images,  respectively. Each image is annotated with 5 ground-truth captions. For both datasets, we followed the splits provided by~\cite{karpathy2015deep}, and constructed reference image groups within the training, validation, and test splits. For completeness, we also reported results on Wang~\etal~\cite{wang2021group}'s dataset for Ref-DIC.

\subsection{Evaluation Metrics}
We applied two kinds of metrics to evaluate the accuracy and distinctiveness of generated captions. For accuracy evaluation, we calculated four commonly used evaluation metrics: BLEU-N (B-N) (1- to 4-grams)~\cite{papineni2002bleu}, ROUGE-L (R)~\cite{lin2004rouge}, METEOR (M)~\cite{banerjee2005meteor}, and CIDEr (C)~\cite{vedantam2015cider}. For distinctiveness evaluation, we developed a new metric named \textbf{DisCIDEr} (DisC). We introduce the DisCIDEr below.

\noindent\textbf{DisCIDEr.} All existing metrics designed for Ref-DIC task fail to fully explore the distinctiveness of each individual n-gram in GT captions of target image. To solve this, we assign these n-grams with different weights according to group-level distinctiveness: if an n-gram occurs frequently in ground-truch captions of reference images, it is less distinctive. As an example in Figure~\ref{fig_5_discider_example}, both target and reference images are describing ``\texttt{red sofa}'', so we should assign lower weights to the word ``\texttt{red}'' in ground-truth captions of $I_t$ at evaluation time. Instead, since only image $I_t$ contains the object ``\texttt{fireplace}'', we should put more weight on it. 

To realize this intuition, we modifies the n-gram weighting procedure of CIDEr by adding a re-weight term. For a target image $I_t$, we denote its generated and ground-truth captions as $c$ and  $S_t=\{s_{t}^i\}_{i=1}^M$, respectively. Similarly, we denote ground-truth captions for reference images $\mathcal{I}_r$ as $S_r=\{s_{r}^j\}_{j=1}^M,r=\{1\ldots K\}$. The number of times an n-gram $\omega_d$ occurs in $s_{t}^j$ is denoted as $h_d(s_{t}^j)$. We modify CIDEr by adding an \textbf{Inverse reference frequency} term after it when calculating $g_d(s_{t}^j)$ for the n-grams in ground-truth captions of $I_t$:
\begin{small}
\begin{align}
    g_d(s_{t}^i) &=  
    \frac{h_d(s_{t}^i)}{\sum_{w_l\in\Omega} h_l(s_{t}^i)}   
    \underbrace{
    \log (\frac{|I|}{\max (1, \sum_{I_p \in I} \min(1,\sum_{s_p^q  \in I_p} h_d(s_{p}^q)))}) }_{\text{Inverse document frequency}}\notag \\
    & {\underbrace{
    \log (\frac{m + K}{n+\sum_{S_u \in {S_{1:K}}} \min (1,\sum_{s_{u}^v\in S_u} h_d(s_{u}^v))})   
    }_{\textcolor{mygreen}{\text{Inverse reference frequency}}}},
\end{align}
\end{small}
where $\Omega$ is the vocabulary of all n-grams and $I$ is the set of all images. $m$ and $n$ are two parameters. In this way, DisCIDEr can evaluate the group-level distinctiveness while preserve advantage of n-gram based metric. We refer reader to~\cite{vedantam2015cider} for more details.
\addtolength{\tabcolsep}{-1pt}
\begin{table}[t]

    \caption{ Comparison of captions accuracy on COCO family with state-of-the-art image captioning models. }
    \centering
  
    \begin{tabular}{lcccccc}
    \hline
    Model & B-1 & B-4 & M & R & C & DisC \\
    \hline    
        \multicolumn{7}{l}{\textbf{Dataset: MS-COCO}} \\
        UpDown~\cite{anderson2018bottom}  & 79.8  & 36.3  & 27.7  & 56.9  & 120.1 & \multicolumn{1}{c}{---}  \\
        AoANet~\cite{huang2019attention}  & 80.2  & 38.9  & 29.2  & 58.8  & 129.8 & \multicolumn{1}{c}{---}  \\
        Transformer~\cite{vaswani2017attention} & 80.0 & 38.2  & 28.9 & 58.2 & 127.3 & 98.7 \\
        $M^2$ Transformer~\cite{2020Meshed}   & 80.8  & 39.1  & 29.2  & 58.6  & 131.2 & \multicolumn{1}{c}{---}  \\
    \cdashline{1-7}[1pt/1pt]
        DiscCap~\cite{luo2018discriminability} & \multicolumn{1}{c}{---}   & 36.1  & 27.4  & 57.3  & 114.3 & \multicolumn{1}{c}{---} \\
        CIDErBtwCap~\cite{wang2020compare} &  \multicolumn{1}{c}{---}   & 38.5  & 29.1  & 58.8  & 127.8 & \multicolumn{1}{c}{---} \\
    \hline
        \multicolumn{7}{l}{\textbf{Dataset: COCO-DIC}} \\
        GdisCap~\cite{wang2021group} & 80.0 & 37.3 & 28.4 & 57.5 & 125.8 & 96.6  \\
        CAGC~\cite{li2020context}  & 80.7 & 38.1 & 28.7 & 57.9 & 127.9 & 98.0 \\
        \textbf{TransDIC (Ours)} & \textbf{81.6} & \textbf{39.3} & \textbf{29.2} & \textbf{58.5} & \textbf{132.0} &\textbf{102.2}\\
    \hline
        \multicolumn{7}{l}{\textbf{Dataset: Wang~\etal~\cite{wang2021group}}} \\
        GdisCap~\cite{wang2021group}  &  80.2 & 37.7  &  28.3 &   57.3 & 126.6 & 97.7 \\
        CAGC~\cite{li2020context}  &  80.4 & 37.7  & 28.7  & 57.6   &  127.2 & 98.0 \\
        \textbf{TransDIC (Ours)} & \textbf{81.0} & \textbf{38.8} & \textbf{29.1} & \textbf{58.2} & \textbf{130.8} &\textbf{101.9}\\
    \hline
    
    \end{tabular}%
    
     \label{table:result_on_coco}
\end{table}%
\addtolength{\tabcolsep}{1pt}
\subsection{Implementation Details}
Following~\cite{anderson2018bottom}, we used the region proposal features extracted by the Faster R-CNN~\cite{ren2016faster} with dimension 2048, and the memory space was of dimension 512. The number of self-attention blocks in Target flow, select and fuse layers in Target-Reference flow were set to 3. In the two-stage matching procedure, the size of $\mathcal{I}_c$ was 500. For $\mathcal{I}_r$, we used images with top-p to top-(p+K-1) highest similarity scores, where p is an adjustable parameter for group similarity and both benchmarks set p to 3, K to 5. Parameters $m$ and $n$ in DisCIDEr were set to 0.8 and 5.0.


\subsection{Comparison with State-of-the-Art Methods}
We reported our results on two kinds of datasets: 1) Our constructed COCO-DIC and Flickr30K-DIC datasets. 2) The dataset proposed in~\cite{wang2021group}. We compared our TransDIC model with three kinds of state-of-the-art models: 1) \textbf{NIC}~\cite{vinyals2015show}, \textbf{Xu~\etal}~\cite{xu2015show}, \textbf{UpDown}~\cite{anderson2018bottom}, 
\textbf{AoANet}~\cite{huang2019attention}, 
\textbf{Transformer}~\cite{vaswani2017attention}, \textbf{$M^2$ Transformer}~\cite{2020Meshed}. They only aim to generate captions with high accuracy. 2) \textbf{DiscCap}~\cite{luo2018discriminability}, \textbf{CIDErBtwCap}~\cite{wang2020compare}. They are designed for the Single-DIC. 3) \textbf{GdisCap}~\cite{wang2021group} that is designed for the Ref-DIC. We also compared \textbf{CAGC}~\cite{li2020context} which use multiple images as input.


\noindent\textbf{Results on COCO-family Benchmarks.} From Table~\ref{table:result_on_coco}, we can observe: 1) For accuracy evaluation, our proposed TransDIC achieves the best performance on all conventional metrics at both COCO-DIC and Wang~\etal~\cite{wang2021group} (\eg, 132.0 vs. 125.8 in GdisCap on CIDEr). Meanwhile, our model outperforms some strong state-of-the-art models (\eg, 132.0 vs. 131.2 in $M^2$ Transformer on CIDEr) in terms of accuracy-based metrics. 2) For distinctiveness evaluation, our model gets the highest scores on DisCIDEr in the two datasets. 

\noindent\textbf{Results on Flickr30K-family Benchmarks.} From Table~\ref{table:result_on_flickr}, we can observe: 1) For accuracy evaluation, our TransDIC achieves the largest performance gains on most metrics while it is narrowly defeated by GdisCap on CIDEr (65.1 vs. 65.6). 2) For distinctiveness-based metrics, our model outperforms GdisCap by 0.2 on DisCIDEr (41.4 vs. 41.2), despite it having a lower CIDEr score.

\begin{table}[t]
    \caption{ Comparison of captions accuracy on Flickr30K family with state-of-the-art image captioning models. }
    \centering
  
    \begin{tabular}{{p{6em}cccccc}}
    \hline
    Model & B-1 & B-4 & M & R & C & DisC\\
    \hline    
        \multicolumn{7}{l}{\textbf{Dataset: Flickr30K}} \\
        NIC~\cite{vinyals2015show} & 66.3  & 18.3 & --- & --- & --- & ---  \\
        Xu~\etal~\cite{xu2015show} & 66.9 & 19.9 & 18.5 & --- & --- & ---  \\
        Transformer~\cite{vaswani2017attention} & 70.7 & 27.7  & 21.4 & 49.0 & 61.2 &39.1  \\
    \hline
        \multicolumn{7}{l}{\textbf{Dataset: Flickr30K-DIC}} \\
        GdisCap~\cite{wang2021group} & 71.7 & 29.0 & 22.1 & 49.6 & \textbf{65.6} & 41.2 \\
        CAGC~\cite{li2020context}  & 72.9 & 29.1 & 21.9 & 50.1 & 62.2 &39.0  \\
        \textbf{TransDIC } & \textbf{73.2} & \textbf{30.1} & \textbf{22.5} & \textbf{50.3} & 65.1 &\textbf{41.4}\\
    \hline
    \end{tabular}%

    \label{table:result_on_flickr}
\end{table}%

\addtolength{\tabcolsep}{1pt}
\begin{table}[t]
  \centering
    \caption{Ablation study of \flows~on COCO-DIC. ``Fuse'' and ``Select'' denote the fuse layer and select layer in the Target-Reference flow, respectively.}

    \begin{tabular}{cc | cccccc}
    \hline
    Fuse & Select & \multicolumn{1}{c}{B-1} & \multicolumn{1}{c}{B-4} & \multicolumn{1}{c}{M} & \multicolumn{1}{c}{R} & \multicolumn{1}{c}{C} & \multicolumn{1}{c}{DisC} \\
    \hline
    \ding{55} & \ding{55} & 80.0    & 38.2  & 28.9 & 58.2 & 127.3 & 98.7 \\
    \ding{55} & \ding{51} & 81.3  & 38.4  & 29.0    & 58.1  & 130.0   & 100.2 \\
    \ding{51} & \ding{55} &  \textbf{81.6}   & 39.1  & \textbf{29.1}  & 58.4  & 131.3 & 101.6  \\
    \ding{51} & \ding{51}  & \textbf{81.6}  & \textbf{39.3}  & \textbf{29.1}  & \textbf{58.5}  & \textbf{132.0}   & \textbf{102.2} \\
    \hline
    \end{tabular}%

    \label{table:abal_module}
\end{table}%
\addtolength{\tabcolsep}{-1pt}

\subsection{Ablation Studies}
We conducted extensive experiments to verify the influences of the proposed \flows~module and group similarity.

\subsubsection{Influence of \flows}
To measure the influence of each component in our proposed Target-Reference flow, we trained an ordinary transformer as the baseline and three variants of our model. 1) Target-Reference flow only contains the select layer: stacked select layers always take the original reference features as input. 2) Target-Reference flow only contains the fuse layer: outputs of the last fuse layer are directly used as the outputs of the Target-Reference flow. 3) Transformer with complete select and fuse layers. All these models were trained on COCO-DIC and the results were shown in Table~\ref{table:abal_module}. 

\noindent \textbf{Results}. As can be observed in rows 2 and 3, two additional components can improve captioning performance consistently in terms of both accuracy and distinctiveness. Above all, our complete model achieves the most promising performance in all metrics.

\subsubsection{Influence of Group Similarity}
To quantify the influence of group similarity, we used different p when choosing K most similar images from $\mathcal{I}_f$, \eg, top-2 to top-6 \textbf{(top2-6)} group when setting p to 2. The results were reported in Table~\ref{table:abal_group}. 

\noindent \textbf{Results}. From Table~\ref{table:abal_group} (a), we can observe: Top1-5 group is surpassed by top3-7 group on DisCIDEr (101.4 vs. 102.2), despite it having a marginal improvement on CIDEr (132.2 vs. 132.0). The results indicate that the most similar reference group is not always helpful to group-level distinctiveness. 
\begin{figure*}[t]
\centering
\includegraphics[width=\linewidth]{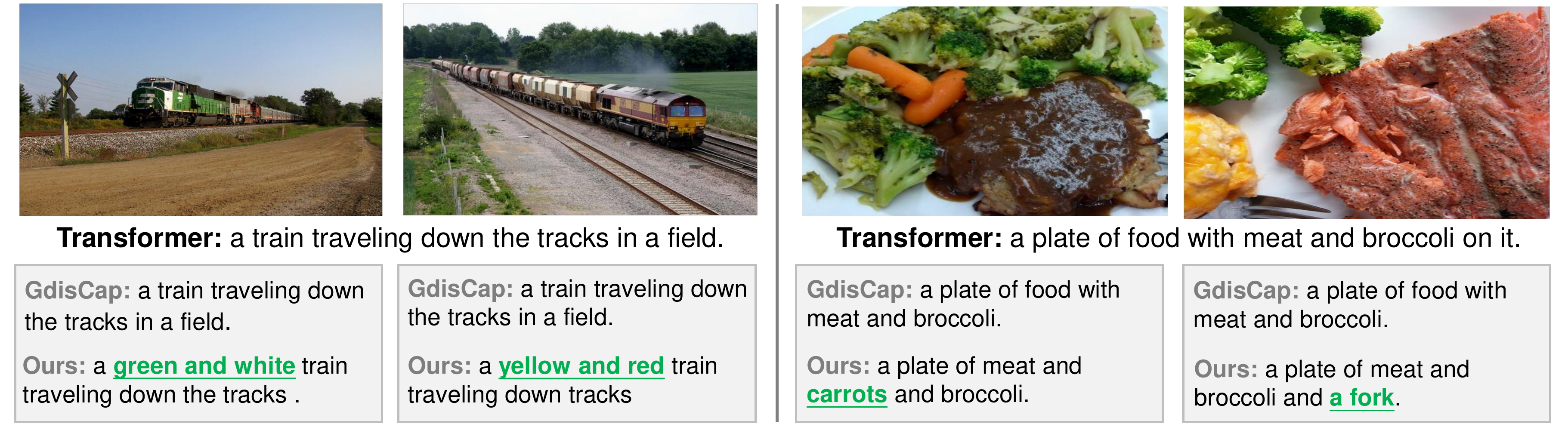}
\caption{Examples of generated captions for two similar images.  The \textcolor{mygreen}{green} words indicate the unique details in the images.}
\label{fig_6_example}
\end{figure*}
\begin{table*}[t]
\caption{Ablation study of group similarity on COCO-DIC and Flickr30K-DIC.}
\subfloat[Comparison of different group trained with \model~on COCO-DIC]{
        \begin{tabular}{c|cccc}
        \hline
        \multicolumn{5}{c}{TransDIC at COCO-DIC}\\
        \hline
        Group & B-1 & B-4 & C & DisC \\[.1em]
        \hline
        top1-5 & \textbf{81.6} & 39.1 & \textbf{132.2} & 101.4\\
        top2-6 & 81.4 & \textbf{39.3} & 131.5 & 101.4\\
        top3-7 & \textbf{81.6} & \textbf{39.3} & 132.0 & \textbf{102.2} \\
        top4-8 & 81.0 & 38.7 & 130.7 & 101.5 \\
        \hline
        \end{tabular}}\hspace{5mm}
        \subfloat[Comparison of different group trained with GdisCap~on COCO-DIC]{
        \begin{tabular}{c|cccc}
        \hline
        \multicolumn{5}{c}{GdisCap at COCO-DIC}\\
        \hline
        Group & B-1 & B-4 & C & DisC \\[.1em]
        \hline
        top1-5 & \textbf{80.4} & \textbf{38.0} & \textbf{126.9} & 97.0\\
        top2-6 & 79.8 & 37.5 & 125.1 & 96.2\\
        top3-7 & 80.0 & 37.3 & 125.8 & 96.6 \\
        top4-8 & 80.2 & 37.6 & 125.3 & \textbf{97.1} \\
        \hline
        \end{tabular}}\hspace{5mm}
\subfloat[Comparison of different group trained with \model~on Flickr30K-DIC]{
        \begin{tabular}{c|cccc}
        \hline
        \multicolumn{5}{c}{TransDIC at Flickr30K-DIC}\\
        \hline
        Metric & B-1 & B-4 & C & DisC \\[.1em]
        \hline
        top1-5 & 72.5 & 29.4 & 64.5 & 40.6\\
        top2-6 & 70.8 & 28.8 & 62.4 & 39.3\\
        top3-7 & 73.2 & 30.1 & 65.1 & 41.6 \\
        top4-8 & \textbf{73.5} & \textbf{30.9} & \textbf{66.7} & \textbf{42.0}\\
        \hline
        \end{tabular}}\hspace{5mm}
\label{table:abal_group}

\end{table*}

\subsection{Qualitative Results}
We illustrated the qualitative results of our proposed \model~and compared it with the Transformer and SOTA Ref-DIC model GdisCap~\cite{wang2021group} in Figure~\ref{fig_6_example}. Naive captioning models generate identical captions for similar images. In contrast, our \model~can describe the unique attributes and objects in the target image. For unique attributes, as shown in Figure~\ref{fig_6_example} (left), TransDIC precisely captures the unique attributes ``\texttt{green and white}'' and ``\texttt{yellow and red}'' for the two trains, respectively. In Figure~\ref{fig_6_example} (right), for unique objects, TransDIC captures unique objects ``\texttt{carrots}'' and ``\texttt{a fork}'' for each individual image. The results demonstrate that TransDIC can generate distinctive captions in terms of unique objects and attributes.

\section{LIMITATIONS}
One possible limitation of our work is that if original human-annotated captions omit some objects or attributes, it will lead to: 1) our proposed two-stage matching mechanism may fail to collect object-/attribute- level similarity reference images. 2) our proposed DisCIDEr may degrade to existing CIDEr. We believe this omitting problem is due to the natural defect of datasets (COCO/Flickr30K): Since these datasets are annotated for general captioning task, human annotators may tend to simply describe the objects while ignoring its attributes (\eg, color) when there is no reference image.

    

\section{Conclusions}
In this paper, we argued that all existing DIC works fail to achieve group-level distinctiveness. To solve this problem, firstly, we introduced a two-stage matching mechanism and proposed two new benchmarks for Ref-DIC. Then, we developed a Transformer-based model for Ref-DIC. Finally, we came up with a new evaluation metric termed DisCIDEr. We conducted extensive experiments to verify the effectiveness of \model. Moving forward, we are going to 1) extend our Ref-DIC into video domains, or 2) design stronger Ref-DIC models.

\begin{acks}
This work was supported by the National Key Research \& Development Project of China (2021ZD0110700), the National Natural Science Foundation of China (U19B2043, 61976185), Zhejiang Natural Science Foundation (LR19F020002), Zhejiang Innovation Foundation(2019R52002), and the Fundamental Research Funds for the Central Universities(226-2022-00087).
\end{acks}
\bibliographystyle{ACM-Reference-Format}
\balance
\bibliography{mm}

\appendix
\onecolumn


\section{User Study Interface}\label{sec:a}

\begin{figure*}[h]
\centering
\includegraphics[width=0.9\linewidth]{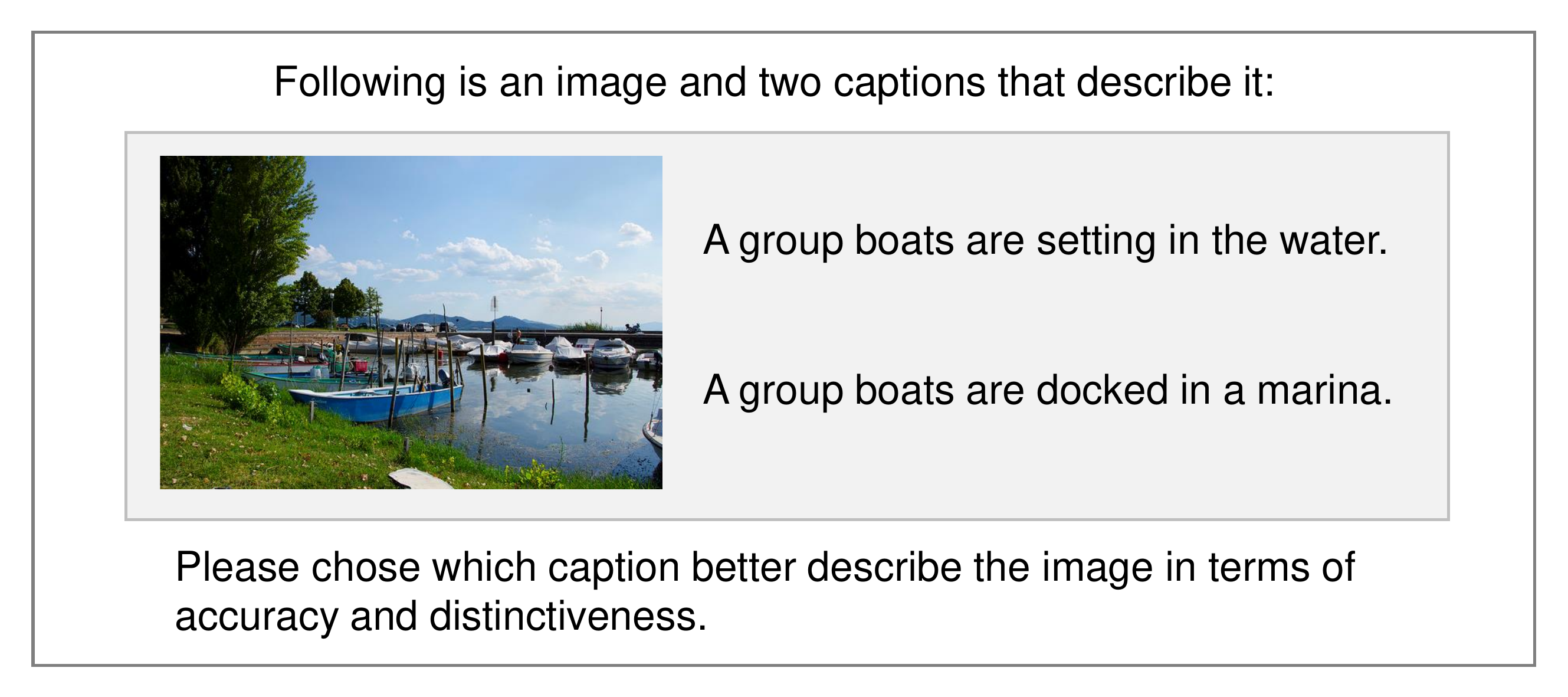}

\caption{We display an image and two captions generated by Transformer~\cite{vaswani2017attention} and our proposed~\model, respectively.  The users are asked to choose which caption better describe the image in terms of
accuracy and distinctiveness.}
\label{fig_s1}
\end{figure*}

We conduct user studies to validate the effectiveness of our proposed \textbf{DisCIDEr}. Specifically, we invite 5 experts and give them a picture and two captions that describe it. They are asked to choose which one better matches the image in terms of accuracy and distinctiveness and the user interface for this is shown in Fig.~\ref{fig_s1}. The caption which got more than 3 votes is regarded as human judgment. We randomly select 100 images from the test split of MS-COCO and ask the experts to give their judgments. A metric agrees with human judgment only if it gives a higher score to the same caption that experts choose. We use the agreement counts between human judgment and the metric as the effectiveness measurement.

\noindent\textbf{User Studies on DisCIDEr.}
We conducted a user study to validate the effectiveness of \textbf{DisCIDEr} with 5 experts. We randomly selected 100 images (100 trials) from the test set. In each trial, an image and two captions (generated by Transformer and TransDIC, respectively) were displayed and these experts are asked to choose the better caption in terms of distinctiveness and accuracy. These captions which got more than 3 votes were regarded as human judgment. We calculated the agreements between human judgments and different metrics (\ie, whether humans and metrics give higher scores to a same caption). Results were reported in Table~\ref{table:user_study}. From Table~\ref{table:user_study}, we can observe that DisCIDEr achieves better agreement than both the accuracy-based metric CIDEr and the distinctiveness-based metric DisWordRate.

\begin{table}[h]
  \centering
    \caption{User study of different metrics}

    \begin{tabular}{c | ccc}
    \hline
    Metric & CIDEr~\cite{vedantam2015cider} & DisWordRate~\cite{wang2021group} & DisCIDEr  \\
    \hline
    Agreements & 58  & 62  & 64  \\
    \hline
    \end{tabular}%
    
    \label{table:user_study}
\end{table}%

\section{Details of these Compared Baselines}\label{sec:c}
In this section, we describe implementation details of two compared state-of-the-art baselines: GdisCap~\cite{wang2021group} and CAGC~\cite{li2020context}. These two group-based captioning baselines are both built on top of the widely-used Transformer~\cite{vaswani2017attention} architecture, which takes a sequence of visual features as input and encode them through multiple self-attention layers. We introduce the two baselines in detail.

\begin{enumerate}[leftmargin=2em]
    \item~\textbf{GdisCap}~\cite{wang2021group}: It develops a Group-based Memory Attention (GMA) module which assigns higher weights to the distinctive regions and two special losses: DisLoss and MemCls loss, which directly encourage the model to generate distinctive words. We re-implement their proposed GMA module and DisLoss.
    
    \item~\textbf{CAGC}~\cite{li2020context}: It is designed to describe a group of target images in the context of another group of related reference images. To this end, they extract visual features for all images in target and reference groups from the ResNet50 network~\cite{he2016deep} (after pool5 layer). Then, these Visual features of images are sent into multiple self-attention layers to enable the interactions within each group (target image interacts with other images only within target group, and same for reference group). Finally, they construct group representations and contrastive representations for both groups and send them into a decoder for captions generation. To make a fair comparison, we replace the visual features with mean pooled bottom-up features~\cite{anderson2018bottom} and re-implement their visual content encoder.

\end{enumerate}

\section{MORE QUALITATIVE RESULTS} \label{sec:d}
\begin{figure*}[b]
\centering
\includegraphics[width=0.95\linewidth]{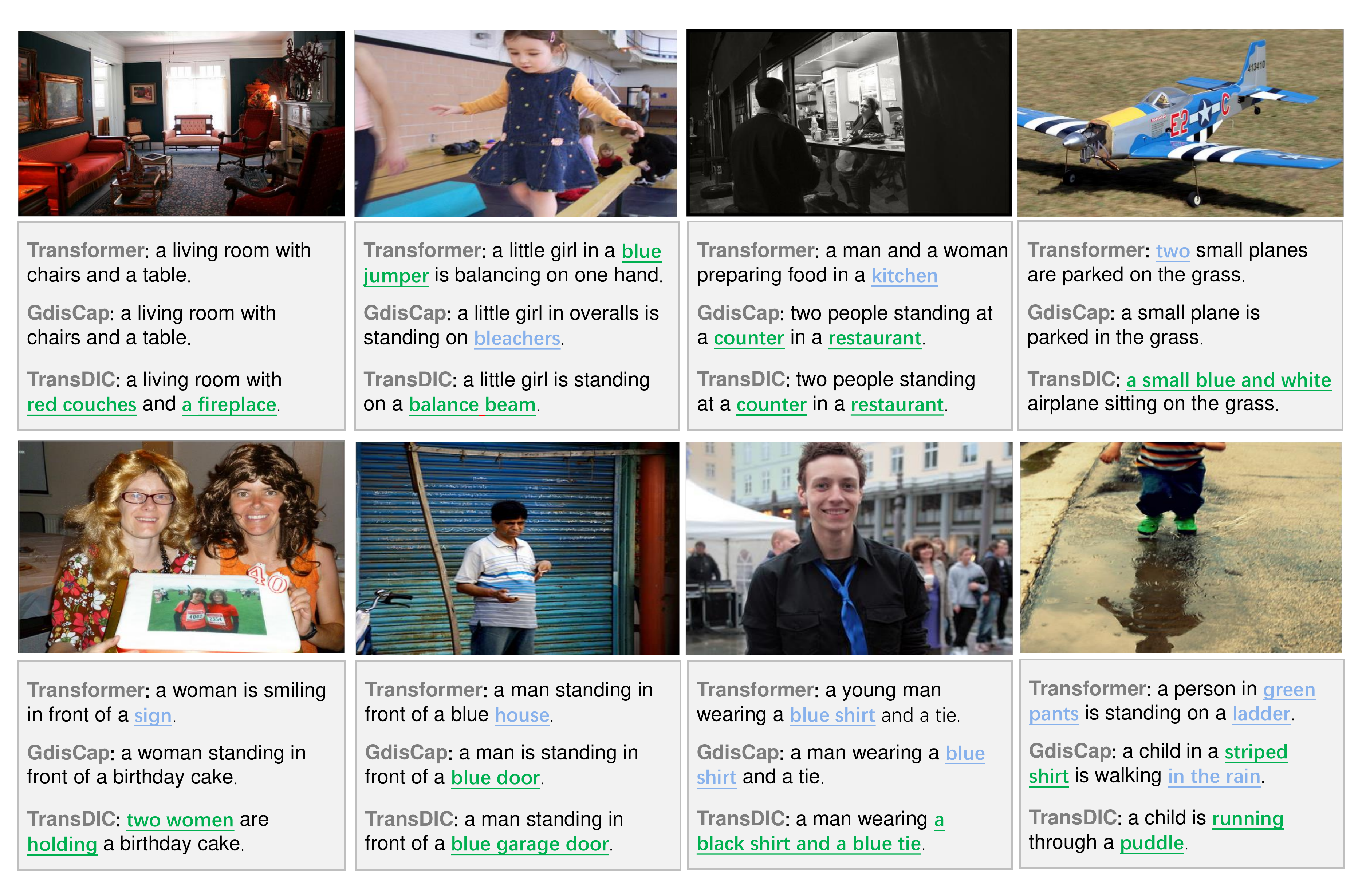}
\caption{Examples of generated captions for two similar images using Transformer, GdisCap and \model. The \textcolor{mygreen}{green} words indicate the unique details in the images while \textcolor{myblue}{blue} denote the mistakes in the generated captions.}
\label{fig_s2}
\end{figure*}

\begin{figure*}[b]
\centering
\includegraphics[width=0.95\linewidth]{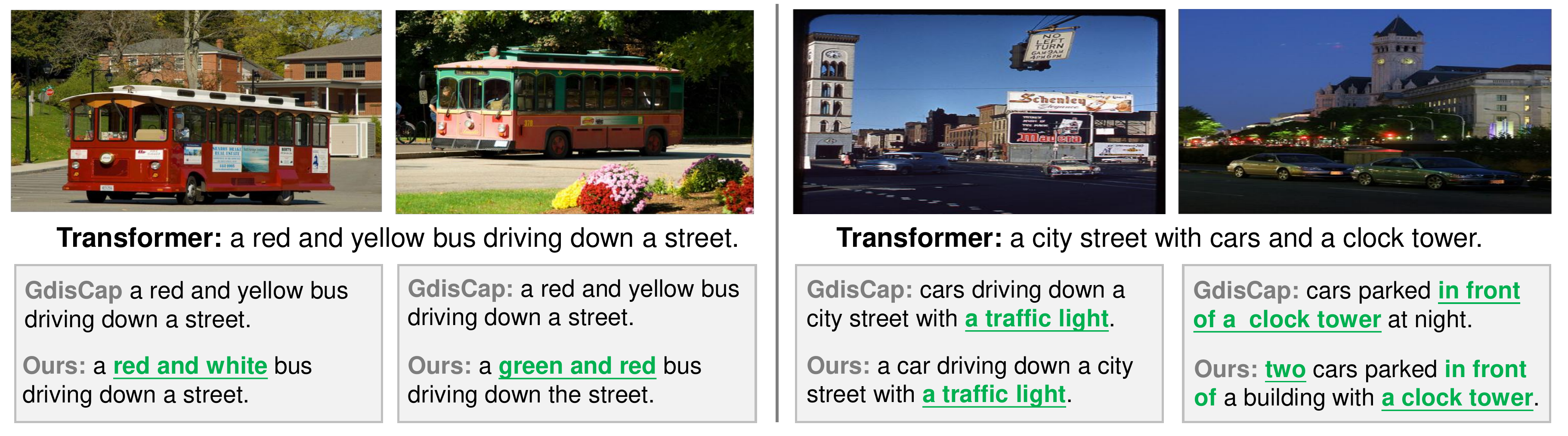}
\caption{Examples of generated captions for two similar images. The \textcolor{mygreen}{green} words indicate the unique details in the images.}
\label{fig_s3}
\end{figure*}

We report more qualitative results in Fig.~\ref{fig_s2} and Fig.~\ref{fig_s3}. In Fig.~\ref{fig_s2}, we compare our \model~with Transformer and GdisCap. Remarkably, our model can precisely capture the unique objects and attributes in the image. For example in Fig.~\ref{fig_s2} (row2, column 3), both Transformer and GdisCap wrongly mix the tie and shirt into “blue
shirt” while \model~correctly matches the “shirt” with “black” and “tie” with “blue”. In terms of numerals, in Fig.~\ref{fig_s2} (row2, column 1), only \model~predicts the right number, whereas others mistake it for ``\texttt{one}''. In Fig.~\ref{fig_s3}, we test all models with two similar images, as can be seen on the left side, \model~ correctly captures the ``\texttt{red and white}'' and ``\texttt{green and red}'' for the two buses, respectively.

\end{document}